\documentclass[11pt,journal,twoside,onecolumn,romanappendices]{IEEEtran}

\usepackage{psfrag,epsfig,graphics}
\usepackage{amsmath,amssymb,multirow}
\usepackage{cite,subfigure}

\usepackage{color}  
\def\cred{\textcolor{black}}  

\def\cmag{\textcolor{black}}  
\definecolor{violet}{rgb}{0.5,0,0.5}

\newcommand{\cp}[1]{\ifmmode {\mathcal{#1}}\else ${\mathcal{#1}}$\fi}
\newcommand{\balpha}{\boldsymbol{\alpha}}

\newcommand{\balphaO}{\boldsymbol{\alpha}}
\newcommand{\balphaN}{\boldsymbol{\alpha}}
\newcommand{\balphaS}{\boldsymbol{\alpha}}
\newcommand{\balphaE}{\boldsymbol{\alpha}}

\newcommand{\bD}{\boldsymbol{D}}

\newcommand{\bI}{\boldsymbol{I}}

\newcommand{\bR}{\boldsymbol{R}}

\newcommand{\sS}{\cp{S}}

\newcommand{\bv}{\boldsymbol{v}}

\newcommand{\bvO}{\boldsymbol{v}}
\newcommand{\bvS}{\boldsymbol{v}}
\newcommand{\bx}{\boldsymbol{x}}

\newcommand{\tr}{\text{trace}}

\newcommand{\E}[1]{E\left\{#1\right\}}

\newcommand{\sgn}{\text{sgn}}

\usepackage{tikz}

\begin{document}

\title{ Steady-state performance of non-negative least-mean-square algorithm and its variants}
\author{\small Jie Chen$^{(1)}$, \emph{Member, IEEE}, \;Jos{\'e} Carlos M. Bermudez$^{(2)}$, \emph{Senior Member, IEEE}, \\   C{\'e}dric Richard$^{(1)}$, \emph{Senior Member, IEEE} \\ \vspace{0.5cm}
\small{\linespread{0.2}$^{(1)}$ Universit{\'e} de Nice Sophia-Antipolis, UMR CNRS 6525, Observatoire de la C{\^{o}}te d'Azur \\
Laboratoire Lagrange, Parc Valrose, 06108 Nice cedex 2 - France \\
tel.: +33.4.92.07.63.94 \hspace{0.5cm} \hspace{0.5cm} fax.:
+33.4.92.07.63.21 \\ jie.chen@unice.fr \hspace{0.5cm}
cedric.richard@unice.fr}
\vspace{0.3cm}\\
\small{$^{(2)}$ Federal University of Santa Catarina \\
Department of Electrical Engineering, 88040-900, Florian{\'o}polis, SC - Brazil \\
tel.: +55.48.3721.7719 \hspace{0.5cm} \hspace{0.5cm} fax.:
+55.48.3721.9280 \\ j.bermudez@ieee.org}\vspace{0.3cm}}
\maketitle

\begin{abstract}
\cred{Non-negative least-mean-square (NNLMS) algorithm and its variants have been proposed for online estimation under non-negativity constraints. The transient behavior of the NNLMS, Normalized NNLMS, Exponential NNLMS and Sign-Sign NNLMS algorithms have been studied in \cite{ChenTSP11,ChenNNLMS_variant}. In this technical report, we derive closed-form expressions for the steady-state excess mean-square error (EMSE) for the four algorithms. Simulations results illustrate the accuracy of the theoretical results. This is a complementary material to~\cite{ChenTSP11,ChenNNLMS_variant}.}
\end{abstract}


\newpage

\section{Introduction}

\begin{color}{black}
Non-negativity is one of the most important constraints that can usually be imposed on parameters to estimate. It is often imposed to avoid physically unreasonable solutions and to comply with natural physical characteristics. Non-negativity constraints appear, for example, in deconvolution problems\cite{Plumbley2003,Moussaoui2006,Lin2006}, image processing~\cite{Benvenuto2010,Keshava2002}, audio processing~\cite{cont2007} and  neuroscience~\cite{cichocki2009nonnegative}.  The Non-Negative Least-Mean-Square algorithm (NNLMS)~\cite{ChenTSP11} and its three variants, namely, Normalized NNLMS, Exponential NNLMS and Sign-Sign NNLMS~\cite{ChenNNLMS_variant}, were proposed to adaptively find solutions of a typical Wiener filtering problem under non-negativity constraints. The transient behavior of these algorithms has been studied in~\cite{ChenTSP11,ChenNNLMS_variant}.  Analytical models have been derived for the mean and mean-square behaviors of the adaptive weights.

This technical report complements the work in~\cite{ChenTSP11,ChenNNLMS_variant} by deriving closed form expressions for the steady-state excess mean square error of each of these algorithms.  These expressions cannot be directly obtained from the transient recursions derived in~\cite{ChenTSP11,ChenNNLMS_variant} because the weight updates include nonlinearities on the adaptive weights.  Moreover, they cannot be derived following the conventional energy-conservation relations~\cite{sayed2008adaptive}.  Hence, new analyses are required to understand the steady-state behavior of these algorithms.

In this technical report, we derive accurate models for the steady-state behaviors of NNLMS and its variants using a common analysis framework, with clear physical interpretation of each term in the expressions. Simulations are conducted to validate the theoretical results.  This work is therefore complements the understanding of the behavior of these algorithms, and introduces a new methodology for the study of the steady-state performance for adaptive algorithms.   We recommend that readers refer to~\cite{ChenTSP11,ChenNNLMS_variant} for a more detailed understanding of the algorithms and their transient behavior. Most notations in this report are consistent with this previous work.
\end{color}

\section{Problem formulation and associated algorithms}

Consider an un known system,  with input-output relation characterized by the linear model
\begin{equation}
             \label{eq:linear.model}
              y(n) = \balpha^{*\top} \bx(n) + z(n)
\end{equation}
with $\balpha^* = [\alpha_1^*, \dots, \alpha_N^*]^\top$ an unknown parameter vector, and $\bx(n) = [x(n), x(n-1), \dots, x(n-N+1)]^\top$ the regressor vector \cmag{with correlation matrix $\bR_x$}. The input signal $x(n)$ and the reference signal $y(n)$ are assumed zero-mean stationary. The modeling error $z(n)$ is assumed to be zero-mean and independent of any other signals \cmag{and has variance $\sigma_z^2$}. Due to inherent physical characteristics of the system, non-negativity is imposed on the estimated coefficient vector $\balpha$. We seek to  identify this system by  minimizing the constrained  mean-square error criterion
\begin{equation}
         \label{eq:mse.nng}
         \begin{split}
                    \balpha^o =& \arg\min_{\balpha} \, E\big\{[y(n) - \balpha^\top \, \bx(n)]^2 \big\}  \\
                                          & \text{subject to }    \alpha_i \geq 0, \quad \forall i.
         \end{split}
\end{equation}
In order to solve this problem in an adaptive and online manner, a LMS-type algorithm,  the non-negative least-mean-square (NNLMS) algorithm was derived in~\cite{ChenTSP11} with weight update relation given by
\begin{equation}
           \label{eq:update.O}
            \balpha(n+1) = \balpha(n) + \eta\,  \bD_{\balpha}(n)\, e(n) \, \bx(n)
\end{equation}
where $\bD_{\balpha}(n)$ denotes the diagonal matrix with $i$th diagonal entry $[\bD_{\balpha}(n)]_{ii} = \alpha_i(n)$, \cmag{$\eta$ denotes a fixed positive step size}, and the estimation error $e(n) = y(n) - \balpha^\top(n) \bx(n)$. Several useful variants were derived to improve the NNLMS properties in some sense~\cite{ChenNNLMS_variant}. Normalized algorithm was proposed to reduce the NNLMS performance sensitivity to the input power, with update relation:
 \begin{equation}
             \label{eq:update.N}
             \balpha(n+1) = \balpha(n) + \frac{\eta}{\bx^\top(n)\bx(n)}\, \bD_{\balpha}(n)\, e(n)\,\bx(n),
 \end{equation}
 where a small positive value $\epsilon$ can possibly be added to the denominator to avoid numerical difficulties.  Exponential NNLMS was  proposed to improve the balance of weight convergent rate:
 \begin{equation}
             \label{eq:update.E}
             \balpha(n+1) = \balpha(n) + \eta\,\bD_{\balpha^{(\gamma)}}(n)\,  e(n)  \, \bx(n)
 \end{equation}
 with $i$th component of $\balpha^{(\gamma)}(n)$ defined as $[\balpha^{(\gamma)}(n)]_i = \sgn\{\alpha_i(n)\}|\alpha_i(n)|^\gamma$.  Finally, Sign-Sign NNLMS was proposed to reduce implementation cost in critical real-time applications, with update relation given by:
 \begin{equation}
             \label{eq:update.S}
               \balpha(n+1) = \balpha(n) + \eta\, \bD_{\balpha}(n) \,\sgn(\bx(n)\, e(n)).
\end{equation}
Reminding us that the error $e(n)$ is a function of  estimated weight $\balpha(n)$, the weight correction terms in these algorithms are highly nonlinear functions of $\balpha(n)$. This makes the theoretical analysis very challenging and significantly \cred{different from those of the LMS-based algorithms employed for solving  unconstrained estimation problems.}

\section{Steady-state mean-square performance analysis}
\cred{Define  the weight} error vector $\bv(n)$ as  the difference between the estimated weight vector $\balpha(n)$ and the real system coefficient vector $\balpha^*$, namely
\begin{equation}
        \label{eq:v.def}
        \bv(n)  = \balpha(n) - \balpha^*.
\end{equation}
Assume that  the step size of the algorithm is chosen to be sufficiently small to ensure the convergence in \cred{the mean and mean-square senses}, and denote  the mean weight estimate at steady-state by $\E{\balpha(\infty)}$. The weight error vector~\eqref{eq:v.def} can then be rewritten by:
\begin{equation}
           \label{eq:v.decomp}
            \begin{split}
            \bv(n) = \underbrace{[ \balpha(n) - \E{\balpha(\infty)} ]}_{\bv'(n)} + \underbrace{[\E{\balpha(\infty)}  - \balpha^*]}_{\E{\bv(\infty)}}  \\
             \end{split}
\end{equation}
where we denote the first difference on RHS of~\eqref{eq:v.decomp}  by $\bv'(n)$, which is the weight error vector with respect \cred{to the mean of the converged weights. The second difference on RHS of~\eqref{eq:v.decomp}} is the weight error~\eqref{eq:v.def} at convergence, i.e., $\E{\bv(\infty)}$.

\cred{In the following analyses we employ the conventional independence assumption, namely, that $\bv(n)$ is independent of $\bx(m)$ for all $m\leq n$~\cite{haykin2005}.}

The estimation error at instant $n$ can be expressed via  these weight errors by
\begin{equation}
         \label{eq:e.alt}
         e(n)  = z(n) - \bv'^\top(n)\bx(n) - E\{\bv^\top(\infty)\}\bx(n)
\end{equation}
It can be verified that the excess mean-square error  can be expressed by
\begin{equation}
         \label{eq:EMSE.decomp}
          \begin{split}
               \text{EMSE}(n) &= \E{[\bv^\top(n)\bx(n)]^2}\\
                                          & =\underbrace{\E{[\bx^\top(n)\, \bv'(n)]^2}}_{\text{EMSE}'(n)} +  \underbrace{\tr\Big\{\bR_x \E{\bv(\infty)} E\{\bv^\top(\infty)\} \Big\}}_{\text{EMSE}^\infty}.
           \end{split}
\end{equation}
The steady-state $\text{EMSE}$ is obtained by taking the limiting value \cred{as $n\rightarrow \infty$}.  Since the second term on RHS of~\eqref{eq:EMSE.decomp} is deterministic, it remains to determine the first term $\text{EMSE}'(n)$ in order to evaluate the steady-state EMSE. The advantage of working with $\bv'(n)$ instead of $\bv(n)$ is that \cred{the expected value of $\bv'(n)$  always converges to 0, i.e., $\E{\bv'(\infty)} =0$, which is not true for $\E{\bv(\infty)}$ in the studied constrained optimization problem.}

\cred{The formulation in \eqref{eq:EMSE.decomp} is general enough to study different non-negativity constrained optimization problems.  When the algorithm solution is unbiased} with respect to real system weights $\balpha^*$, the contribution of  $\text{EMSE}^{\infty}$ will be zero.  \cred{When the algorithm solution} is unbiased with respect to the constrained solution $ \balpha^o$, \cred{then $\text{EMSE}^{\infty}$} accounts for the error that is directly generated due to the constraints. Otherwise, $\E{\bv(\infty)}$ can be determined by \cred{running the recursive models derived in~\cite{ChenTSP11,ChenNNLMS_variant} for the mean weight behaviors.}

\cred{For the analyses that follow, we distinguish the weights into two sets:}
\begin{itemize}
       \item   Set  $\sS_+$ denotes the indices of the weights that converge in mean to positive values  at  steady-state, namely, $$\sS_+ = \{i :  E\{\alpha_i(\infty)\} > 0\}.$$
       \item   Set  $\sS_0$ denotes the indices of the weights that converge in mean to zero at  steady-state, namely, $$\sS_0 = \{i :  E\{\alpha_i(\infty)\} = 0\}.$$
\end{itemize}
Considering that the non-negativity constraint is always satisfied at steady-state, \cred{then $E\{\alpha_i(\infty)\} = 0$ implies} that $\alpha_i(\infty) =0$ for $i\in\sS_0$ \cred{for all realizations.} The weight error vector $\bv_{\sS_0}(\infty)$ is then deterministic and given by
\begin{equation}
            \label{eq:const.v1}
            v_{i}(\infty) =  - \alpha^*_{i}, \qquad \text{for }  i\in\sS_0
\end{equation}
and, consequently,
\begin{equation}
             \label{eq:const.v2}
              v_{i}'(\infty) =  0, \qquad \text{for }  i\in\sS_0.
\end{equation}
Now  let ${\overline{\bD}^{-1}_{\balpha}(n)}$ be a diagonal matrix with entries
\begin{equation}
            \label{eq:defD_1}
            [{\overline{\bD}^{-1}_{\balpha}(n)}]_{ii} = \begin{cases}
            \frac{1}{\alpha_i(n)}, \qquad i\in\sS_+  \\
            0,  \qquad \quad\; \;   i\in\sS_0
            \end{cases}
\end{equation}
and $\overline{\bI}$ be the diagonal matrix such that
\begin{equation}
            [\,\overline{\bI} \, ]_{ii} = \begin{cases}
            1, \qquad \quad\; \;  i\in\sS_+  \\
            0,  \qquad \quad\; \;   i\in\sS_0
            \end{cases}
\end{equation}
With these matrices, \cred{we have that}
\begin{equation}
      {\overline{\bD}^{-1}_{\balpha}(n)}  \bD_{\balpha}(n) = \overline{\bI},
\end{equation}
\cred{and, as $n\rightarrow\infty$, }
\begin{equation}
            \E{\bD_{\balpha}(\infty)}  \,  \overline{\bI} =   \E{\bD_{\balpha}(\infty)}.
\end{equation}

With these definitions and notations at hand, \cred{we now perform} the steady-state analysis for non-negative least-mean-square algorithm and its variants.

\subsection{Steady-state performance for NNLMS}
Subtracting $\E{\balpha(\infty)}$ from both sides of~\eqref{eq:update.O}, we have the weight error update relation
\begin{equation}
             \label{eq:vupdate.O}
             \bvO'(n+1) = \bvO' (n)  + \eta\, e(n) \bD_{\balphaO}(n)\,\bx(n).
\end{equation}
Now taking the expected value of the  weighted square-norm $\|\cdot\|_{\overline{\bD}^{-1}_{\balphaO}(n)}^2$, we have
\begin{equation}
     \label{eq:normv.O}
\begin{split}
       E \Big\{  \|\bvO'(n+1)\|^2_{\overline{D}^{-1}_{\balphaO}(n)} \Big\} =&  E\Big\{\| \bvO'(n)+ \eta\,\bD_{\balphaO} (n) \,\bx(n) \, e(n)\|^2_{\overline{D}^{-1}_{\balpha}(n)} \Big\}    \\
      =&  E \Big\{  \|\bvO'(n)\|^2_{\overline{D}^{-1}_{\balphaO}(n)}  \Big\}  + 2 \eta\, E \Big\{\bvO'^\top(n) \overline{\bI} \,  \bx(n)\,e(n)  \Big\}
        + \eta^2\, E \Big\{ \bx^\top(n) \overline{\bI} \, \bD_{\balphaO}(n)  \bx(n) \,e^2(n)  \Big\}.
\end{split}
\end{equation}
\cred{Assuming convergence, we consider the following relation to be valid at steady-state:}
\begin{equation}
             \label{eq:limeq.O}
               \lim_{n\rightarrow \infty}E\{ \|\bvO'(n+1)\|^2_{\overline{\bD}^{-1}_{\balphaO}(n)} \} =  \lim_{n\rightarrow \infty} E\{\|\bvO'(n)\|^2_{\overline{\bD}^{-1}_{\balphaO}(n)}\}.
\end{equation}
The expected value of the second term  on RHS of~\eqref{eq:normv.O} with $n\rightarrow\infty$ is given by
\begin{equation}
        \label{eq:exp2.O}
         \begin{split}
           \lim_{n\rightarrow \infty}  &E \Big\{\bvO'^\top(n) \overline{\bI} \, \bx(n)\,e(n)  \Big\}  \\
           =& \lim_{n\rightarrow \infty}E \Big\{ \bvO'^\top(n) \,\overline{\bI} \, \bx(n) \big[z(n) - \bvO'^\top(n)\bx(n) - E\{\bvO^\top(\infty)\}\,\bx(n) \big]  \Big\} \\
           = & -\lim_{n\rightarrow \infty}E \Big\{ \bvO'^\top(n) \,\overline{\bI} \, \bx(n)\bx^\top(n)   \bvO'(n) +  \bvO'^\top(n) \,\overline{\bI} \, \bx(n)\bx^\top(n) E\{\bvO(\infty)\}  \Big\}  \\
          = & -\lim_{n\rightarrow \infty} E \Big\{ \bvO'^\top(n) \bx(n)\bx^\top(n)   \bvO'(n) \Big\} \\
           = &  -\text{EMSE}'(\infty)
           \end{split}
\end{equation}
where we have considered the property $\E{\bvO'(\infty)} = 0$ and $\bv'^\top(n)\overline{\bI}=\bv'^\top(n)$ due to property~\eqref{eq:const.v2}. The expected value of the third term  \cred{on the RHS} of~\eqref{eq:normv.O}  with $n\rightarrow\infty$ is given by
\begin{equation}
   \label{eq:exp3.O}
         \begin{split}
                \lim_{n\rightarrow\infty}    &E \Big\{ \bx^\top(n) \overline{\bI} \, \bx(n) \, e^2(n)  \Big\}  \\
              =&\lim_{n\rightarrow\infty}     E \Big\{ \bx^\top(n) \overline{\bI} \,\bD_{\balphaO}(n)  \bx(n) \, (z(n) - \bvO'^\top(n)\bx(n) - E\{\bv^\top(\infty)\} \bx(n))^2  \Big\}.
         \end{split}
\end{equation}
We assume that at steady-state $\|\bx(n)\|_{\bD_{\balpha}(n)}^2$ is independent of $e^2(n)$, which is  similar to the  approximation performed in~\cite{sayed2008adaptive}. This expected value can be expressed by
\begin{equation}
        \lim_{n\rightarrow\infty}     E \Big\{ \bx^\top(n) \overline{\bI} \, \bx(n) \,e^2(n)  \Big\}
           \approx \; \tr\{\E{\bD_{\balphaO}(\infty)} \bR_x \} \,( \sigma_z^2 - \text{EMSE}'(\infty) - \text{EMSE}^\infty).
\end{equation}
Now using the relation~\eqref{eq:limeq.O} to~\eqref{eq:exp3.O} in the norm equality~\eqref{eq:normv.O} gives us the relation
\begin{equation}
        - 2\,\eta\, \text{EMSE}'(\infty) +\eta^2 \,\tr\{\E{\bD_{\balphaO}(\infty)} \bR_x \} \,( \sigma_z^2 + \text{EMSE}'(\infty) + \text{EMSE}^\infty) = 0,
\end{equation}
which yields
\begin{equation}
            \text{EMSE}'(\infty) = \frac{\eta \,\sigma_z^2 \, \tr\{\E{\bD_{\balphaO}(\infty)} \bR_x \}} {2 - \eta\, \tr\{\E{\bD_{\balphaO}(\infty)} \bR_x \} }
           + \frac{\eta \,\text{EMSE}^\infty } {2 - \eta\, \tr\{\E{\bD_{\balphaO}(\infty)} \bR_x \} }
\end{equation}
In the above expression, the first term accounts for the EMSE contribution associated with unbiased components, which is equivalent to EMSE of the LMS algorithm with component-wise step sizes $\E{\alpha_i(\infty)}$. This result is reasonable when observing the weight update relation~\eqref{eq:update.O}. The second term accounts for EMSE introduced in the adaptive process by the bias error with respect to unconstrained solution. Finally considering the relation~\eqref{eq:EMSE.decomp}, i.e., adding the direct bias contribution, the excess mean-square error at \cred{steady-state} is given by:
\begin{equation}
             \label{eq:EMSE.O}
             \text{EMSE}(\infty) =     \frac{\eta \,[\sigma_z^2 \, \tr\{\E{\bD_{\balphaO}(\infty)} \bR_x \}  + \text{EMSE}^\infty ]} {2 - \eta\, \tr\{\E{\bD_{\balphaO}(\infty)} \bR_x \} } +\text{EMSE}^\infty
\end{equation}

\subsection{Steady-state performance for Normalized NNLMS}
For systems with large filter length, it is common to neglect  correlation between the denominator $\bx^\top(n)\bx(n)$ and the other terms, since the former tends to vary much slower~\cite{Samson1983,almeida2005}.  \cred{Moreover, for sufficiently large values of $N$, the} Normalized NNLMS can then be approximated by the NNLMS algorithm with the equivalent step size:
\begin{equation}
          \tilde{\eta} = \frac{\eta}{N\,\sigma_x^2}.
\end{equation}
Based on this approximation, the steady-state EMSE for Normalized NNLMS is given directly by using $\tilde{\eta}$ in~\eqref{eq:EMSE.O}:
\begin{equation}
             \label{eq:EMSE.N}
             \text{EMSE}(\infty) =     \frac{ \tilde{\eta} \,[\sigma_z^2 \, \tr\{\E{\bD_{\balphaN}(\infty)} \bR_x \}  + \text{EMSE}^\infty ]} {2 - \tilde{\eta}\, \tr\{\E{\bD_{\balphaN}(\infty)} \bR_x \} } +\text{EMSE}^\infty.
\end{equation}

\subsection{Steady-state performance for Exponential NNLMS}

\cred{Let  ${\overline{\bD}^{-1}_{\balphaE^{(\gamma)}}(n)}$ be a matrix defined with the same structure of~\eqref{eq:defD_1}, with entries $[{\overline{\bD}^{-1}_{\balphaE^{(\gamma)}}(n)}]_{ii} = \frac{1}{\alpha_i^\gamma(n)}$ for $i\in\sS_+$, $[{\overline{\bD}^{-1}_{\balphaE^{(\gamma)}}(n)}]_{ii} =0$ otherwise.}   Following the same steps \cred{that led to the EMSE for the NNLMS algorithm, except by taking the} weighted square-norm $\|\cdot\|_{\overline{\bD}^{-1}_{\balphaE^{(\gamma)}}(n)}^2$ when writing the norm equality~\eqref{eq:normv.O}, yields the following steady-state performance for Exponential NNLMS:
 \begin{equation}
             \label{eq:EMSE.E}
             \text{EMSE}(\infty) =     \frac{{\eta} \,[\sigma_z^2 \, \tr\{\E{\bD_{\balphaE^{(\gamma)}}(\infty)} \bR_x \}  + \text{EMSE}^\infty ]} {2 -{\eta}\, \tr\{\E{\bD_{\balphaE^{(\gamma)}}(\infty)} \bR_x \} } +\text{EMSE}^\infty
\end{equation}

\subsection{Steady-state performance for Sign-Sign NNLMS}
In this subsection, we shall derive EMSE for Sign-Sign NNLMS in detail due to the particular nonlinearity introduced by ${\rm sgn}$ function. Subtracting $\E{\balphaS(\infty)}$ from both sides of the weight update relation~\eqref{eq:update.S}, we have the relation:
\begin{equation}
            \bvS'(n+1) = \bvS'(n) + \eta\, \bD_{\balphaS}(n)\, \sgn(\bx(n)\, e(n))
\end{equation}
Now taking the expected value of the weighted square-norm $\|\cdot\|_{\overline{\bD}^{-1}_{\balphaS}(n)}^2$, we have
\begin{equation}
          \label{eq:normv.SS}
           \begin{split}
            E\Big\{\|\bvS'(n+1)\|^2_{\overline{\bD}_{\balphaS}^{-1}(n)} \Big\} =&  E\Big\{\|  \bvS'(n) + \eta\, \bD_{\balphaS}(n)\, \sgn(\bx(n)\, e(n))  \|^2_{\overline{\bD}_{\balphaS}^{-1}(n)} \Big\} \\
                                       = & E\Big\{\|\bvS'(n)\|^2_{\overline{\bD}_{\balphaS}^{-1}(n)} \Big\}
                                    + \eta^2\, \E{{\rm sgn} (\bx^\top(n)e(n)) \overline{\bI}\, {\bD}_{\balphaS}(n) {\rm sgn} (\bx(n) e(n))} \\
                                     & + 2\,\eta\,\E{ \bvS'^{\top}(n) \overline{\bI} \, {\rm sgn} ( \bx(n)\, e(n))}
           \end{split}
\end{equation}
\cred{Assuming convergence, we consider the following relation to be valid at steady-state:}
\begin{equation}
             \label{eq:limeq.S}
               \lim_{n\rightarrow \infty}E\{ \|\bvS'(n+1)\|^2_{\overline{\bD}^{-1}_{\balphaS}(n)} \} =  \lim_{n\rightarrow \infty} E\{\|\bvS'(n)\|^2_{\overline{\bD}^{-1}_{\balphaS}(n)}\}.
\end{equation}
The expected value of the second term  on RHS of~\eqref{eq:normv.SS} with $n\rightarrow\infty$ is given by
\begin{equation}
       \begin{split}
            \lim_{n\rightarrow \infty} &\E{{\rm sgn} (\bx^\top(n)e(n)) \overline{\bI}\, {\bD}_{\balphaS}(n)     {\rm sgn} (\bx(n) e(n))}\\
       = & \lim_{n\rightarrow \infty} \E{\sum_{i=1}^N {\rm sgn }(x_i^2(n)e^2(n))\alpha_{i}(n)} \\
       = &\, \tr\{\E{\bD_{\balphaS}(\infty)}\}.
       \end{split}
 \end{equation}
 The expected value of the third term on RHS of~\eqref{eq:normv.SS}  with $n\rightarrow\infty$ is given by
\begin{equation}
          \label{eq:exp3.SS}
          \begin{split}
               \lim_{n\rightarrow\infty} &\E{ \bvS'^{\top}(n)    \overline{\bI} \,  {\rm sgn} ( \bx(n)\, e(n))} \\
          = &   \lim_{n\rightarrow\infty}  \E{\bvS'^\top(n)   \, \overline{\bI} \,     \E{\sgn(\bx(n)\,e(n)|\bvS'(n))} } \\
            = & \frac{2}{\pi}  \lim_{n\rightarrow\infty} \E{\bvS'^\top(n)\, \overline{\bI} \, \sin^{-1} \left( -\frac{\bR_x\, \bvS'(n)}{\sigma_x \sigma_{e|\bvS'(n)}} \right)}
          \end{split}
\end{equation}
where we used \cred{Price's} theorem to obtain this result since $x_i(n)$ and $e(n)$ are jointly Gaussian \cred{when conditioned on $\bv'(n)$~\cite{ChenNNLMS_variant}. The variance of $e(n)$ is given by}
\begin{equation}
         \begin{split}
       \sigma_{e|\bvS'(n)}^2 &= \E{ \big[z(n) - \bvS'^\top(n) \bx(n) - E\{\bvS^\top(\infty)\} \bx(n) \big]^2 | \bvS'(n)} \\
                                               & = \sigma_z^2 + \tr\{\bR_x \bvS'(n)\bvS'^\top(n)  \}  + \tr\{\bR_x E\{\bvS(\infty)\} E\{\bvS^\top(\infty)\} \}.
       \end{split}
\end{equation}
The term in the expectation operator in~\eqref{eq:exp3.SS} is highly nonlinear due to function $\sin^{-1}(\cdot)$. It  is reasonable to approximate $\sin^{-1}(\cdot)$  using the linear expansion about the point $\E{\bvS'(\infty)}$, since the weight errors fluctuate around $\E{\bvS'(\infty)}$ at steady-state. With the fact that $\E{\bvS'(\infty)} = 0$, we have
 \begin{equation}
       \begin{split}
             \lim_{n\rightarrow\infty} &\E{\bvS'^\top(n)   \, \overline{\bI} \,   \sin^{-1} \left( -\frac{\bR_x\, \bvS'(n)}{\sigma_x \sigma_{e|\bvS(n)}} \right)}  \\
       \approx&   \lim_{n\rightarrow\infty} \E{  \bvS'^{\top}(n)\, \, \overline{\bI} \,   \  \frac{\bR_x}{\sigma_x\sigma_{e|\E{\bvS'(\infty})}}\bvS'(n)}  \\
       =& - \frac{2}{\pi \, \sigma_x\sigma_{e|\E{\bv'(\infty)}}} \, \text{EMSE}'(\infty)
       \end{split}		
 \end{equation}
 with
 \begin{equation}
           \begin{split}
                          \sigma_{e|\E{\bvS'(\infty)}}^2  = \sigma_z^2 + \text{EMSE}^\infty
           \end{split}
 \end{equation}
\cred{Substituting these results} into the norm equality~\eqref{eq:normv.SS}, we have the equation
 \begin{equation}
             \eta^2 \,  \tr\{\bD_{\balphaS}(\infty)\}    - 2\,\eta\, \frac{2}{\pi \, \sigma_x\sigma_{e|\E{\bv'(\infty)}}} \text{EMSE}'(\infty)  =0
 \end{equation}
 which yields
\begin{equation}
                \text{EMSE}'(\infty) = \frac{\eta\,\pi}{4}  \tr\{\E{\bD_{\balphaS}(\infty)}\} \sigma_x\sigma_{e|\E{\bvS'(\infty)}}
\end{equation}
\cred{Finally, from~\eqref{eq:EMSE.decomp}} the performance for Sign-Sign NNLMS algorithm at steady-state is given by
\begin{equation}
               \text{EMSE}=  \frac{\eta\,\pi}{4}  \tr\{\E{\bD_{\balphaS}(\infty)}\} \sigma_x \sqrt{\sigma_z^2+\text{EMSE}^\infty } + \text{EMSE}^\infty.
\end{equation}

\section{Experiment validation}

\begin{figure*}[!th]
              \subfigure[\; Original NNLMS]{
                    \label{fig:EMSE_NNLMS}
   		 \centering
      		 \includegraphics[trim = 30mm 20mm 30mm 25mm, clip,scale=0.6]{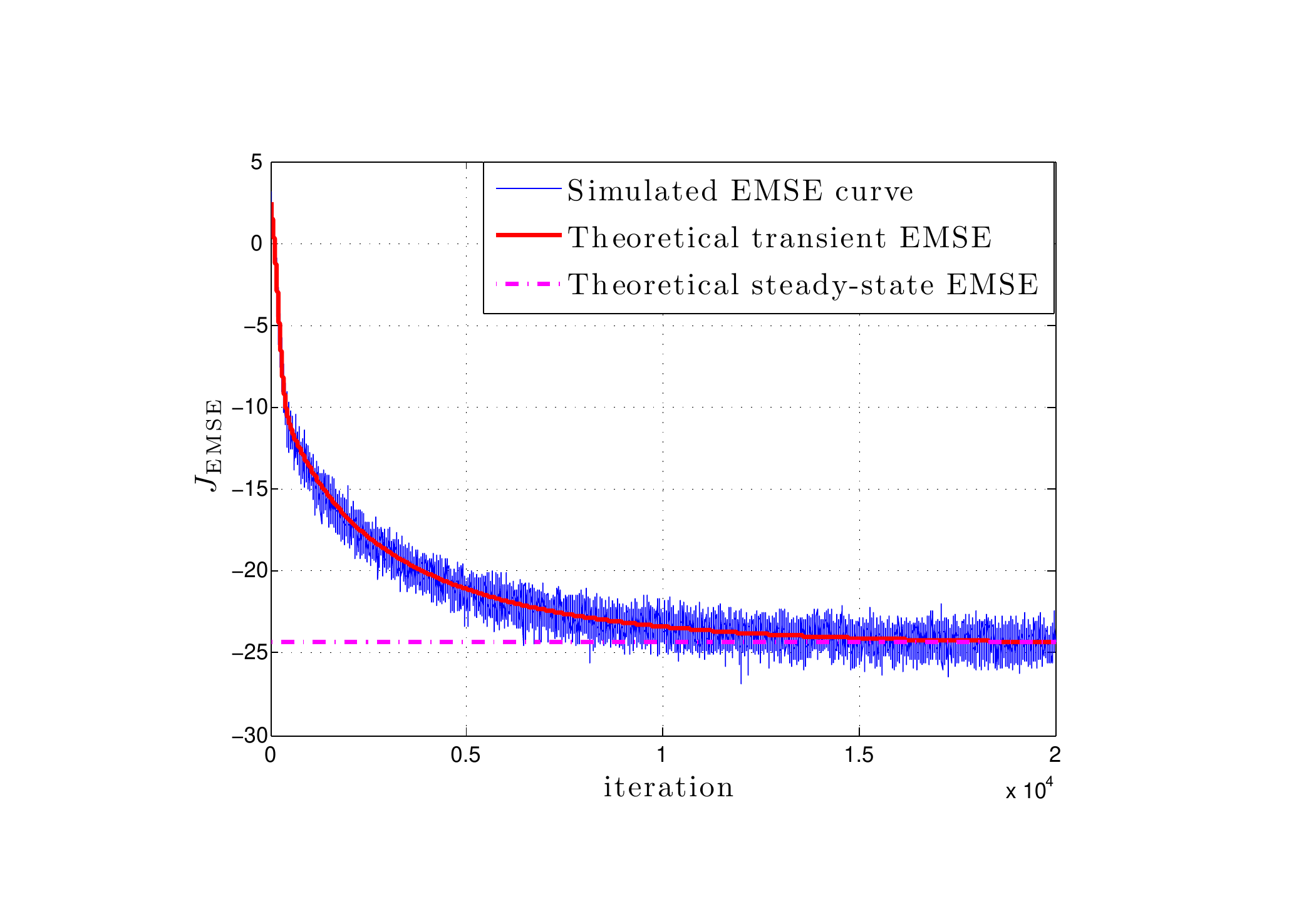}  }
            \subfigure[\; Normalized NNLMS]{
                   \label{fig:EMSE_NNLMS_N}
   		\centering
      		\includegraphics[trim = 30mm 20mm 30mm 25mm, clip,scale=0.6]{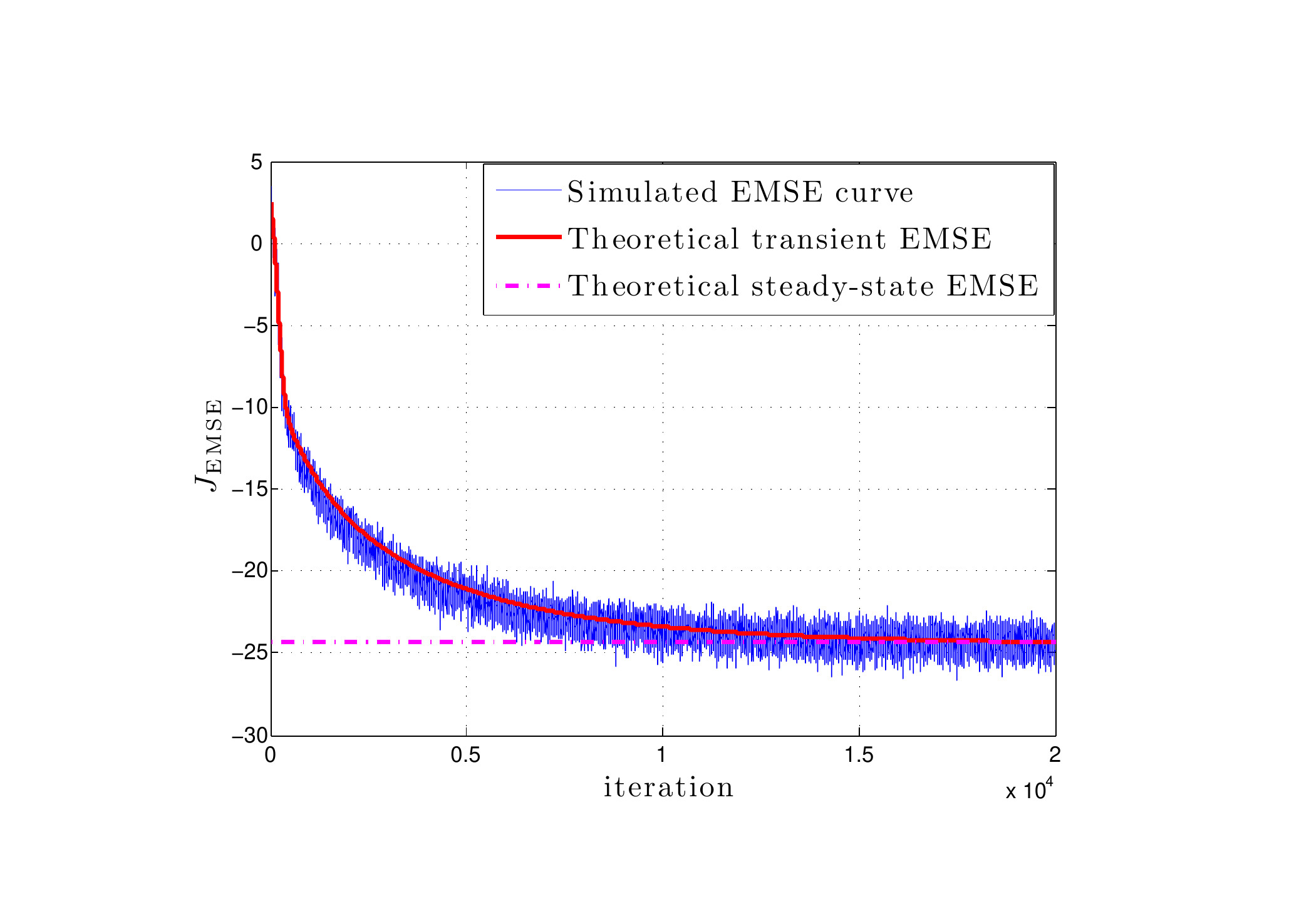}  }
        	  \subfigure[\; Exponential NNLMS]{
                   \label{fig:EMSE_NNLMS_E}
   		\centering
      		\includegraphics[trim = 30mm 20mm 30mm 20mm, clip,scale=0.6]{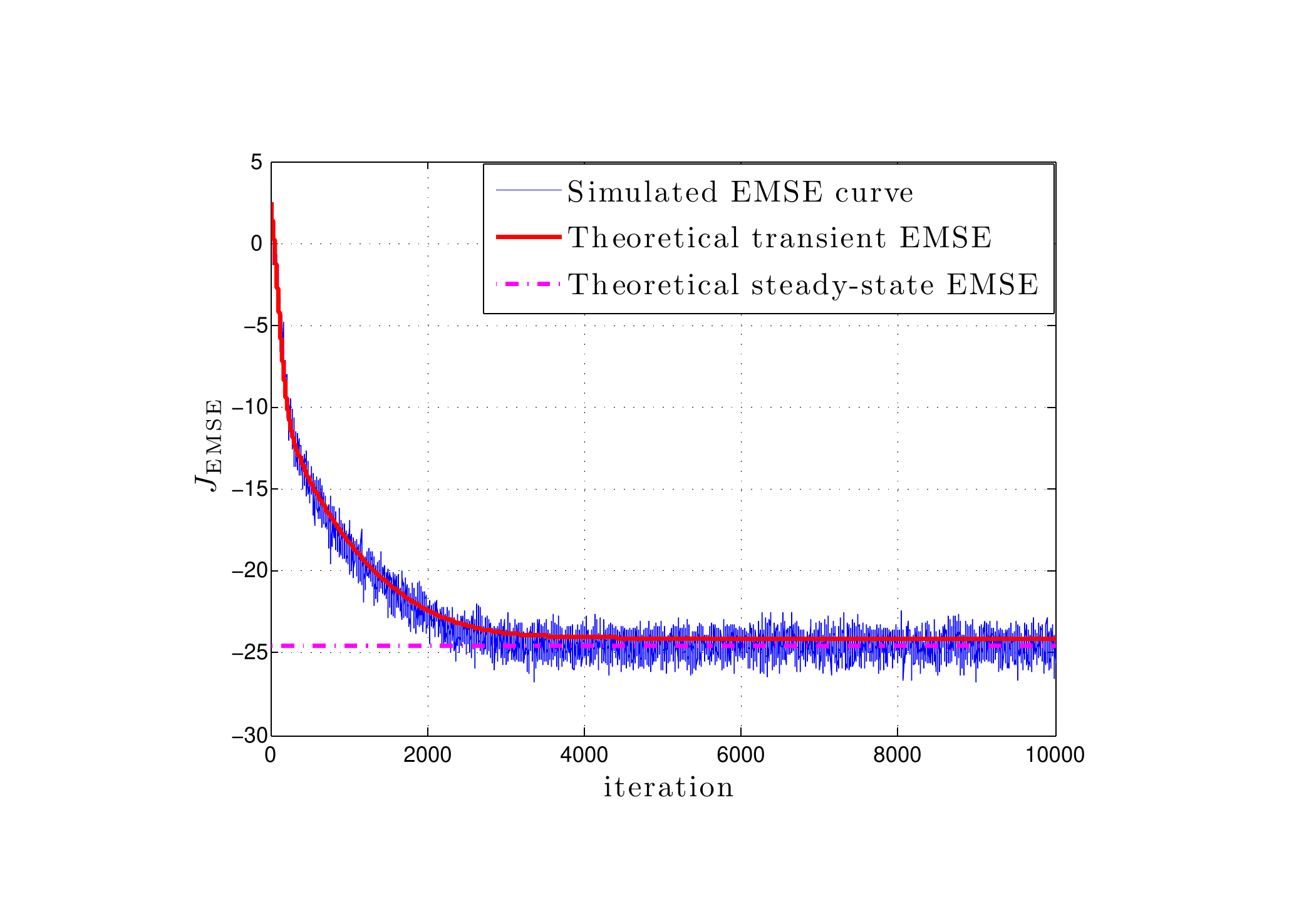}  }
	  \subfigure[\; Sign-Sign NNLMS]{
                   \label{fig:EMSE_NNLMS_SS}
   		\centering
      		\includegraphics[trim = 30mm 20mm 30mm 20mm, clip,scale=0.6]{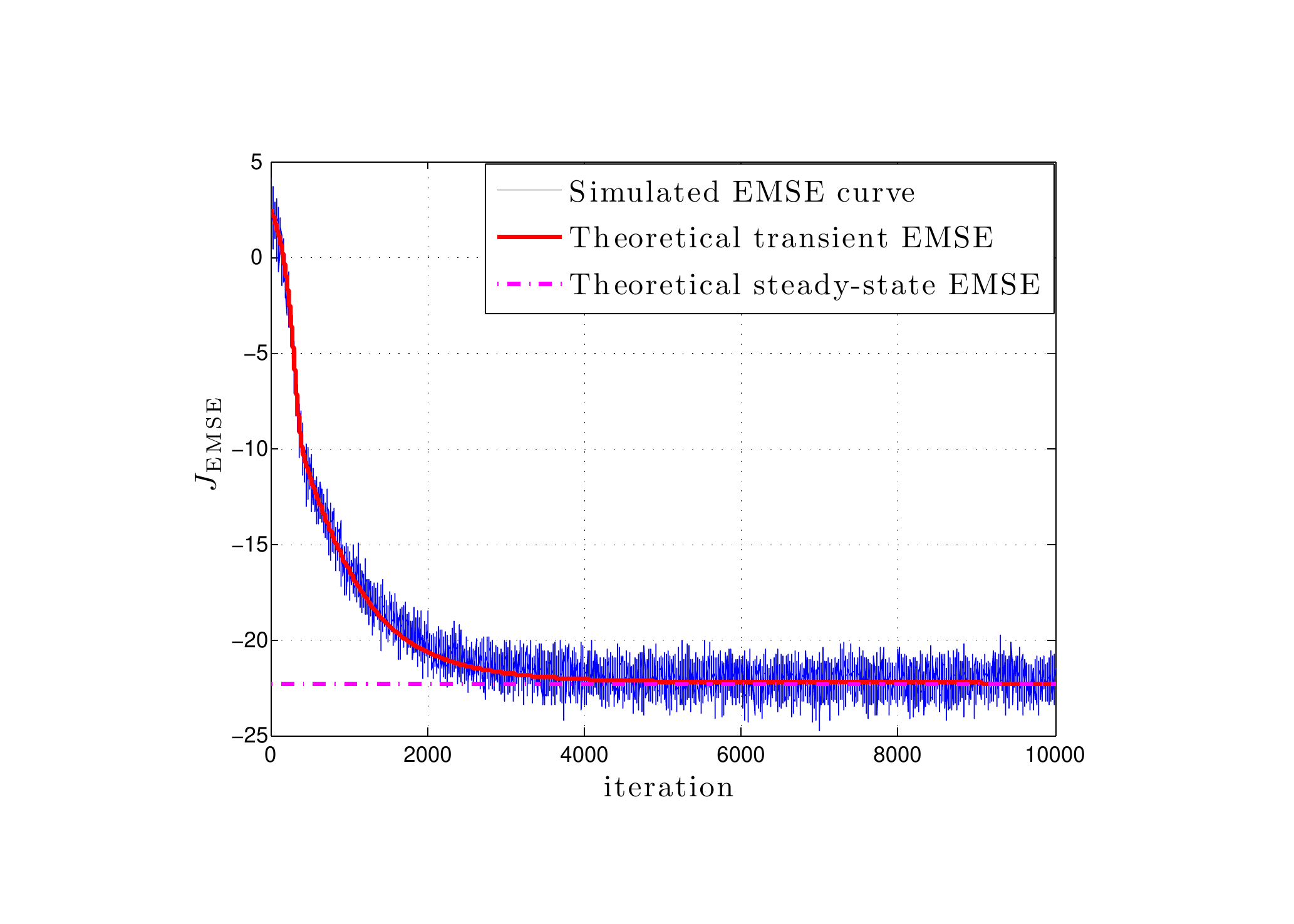} }
	\caption{Steady-state EMSE model validation for NNLMS and its variants.}
	\label{fig:EMSE}
\end{figure*}

In this section, we present examples to illustrate the \cred{correspondence} between theoretical steady-state EMSE and simulated results, for the NNLMS algorithm and its variants. Consider an unknown system of order $N=15$ and weights defined by
\begin{equation}
\begin{split}
           \balpha^* = [&0.8, \, 0.6, \, 0.5, \,  -0.05, \,  0.4, \,-0.04, \, 0.3, \, -0.03, \\
           &0.2, \, -0.02, \, 0.1, \, -0.01, \, 0, \, 0, \,0]^\top,
\end{split}
\end{equation}
where negative coefficients were explicitly included to activate the non-negativity constraint. The input signal was the first-order AR progress given by $x(n) = 0.5\,x(n-1) + w(n)$, where $w(n)$ is an i.i.d. zero-mean Gaussian sequence with variance $\sigma_w^2=0.75$ (so that $\sigma_x^2=1$) and independent of any other signal. The additive independent noise $z(n)$ was zero-mean i.i.d. Gaussian with variance $\sigma_z^2 = 0.01$. The adaptive weights were initialized with $\alpha_i(0)=0.1$ for $i=1,\ldots,N$. The step sizes were equal to $\eta = 0.01N\sigma_x^2$ for NNLMS and $\eta = 0.01$ for the NNLMS, Exponential NNLMS and Sign-Sign NNLMS algorithms. Monte Carlo simulation results were obtained by averaging 100 runs. Figure~\ref{fig:EMSE} shows the simulation results and the behavior predicted by the analytical models. The theoretical transient EMSE behaviors were obtained using results in~\cite{ChenTSP11,ChenNNLMS_variant}, and the theoretical steady-state EMSE (horizontal dashed lines)  were calculated by the expressions derived in this report. These figures clearly validate the proposed theoretical results.

\section{Conclusion}
In this report, we derived closed-form expressions to characterize steady-state excess mean-square errors for the non-negative LMS algorithm and its variants. Experiments illustrated the accuracy of the derived results. Future work may include derive other useful variants of NNLMS and study their stochastic performance.

\bibliographystyle{IEEEbib}
\bibliography{ref}

\end{document}